\pdfoutput=1
\documentclass[10pt]{article}

\usepackage[letterpaper, textwidth=5.5in, textheight=9in, centering,
            headheight=12pt, headsep=25pt, footskip=30pt]{geometry}
\usepackage[T1]{fontenc}
\usepackage[utf8]{inputenc}
\usepackage{times}          
\usepackage{microtype}

\usepackage{amsmath,amssymb,amsthm,mathtools,bm}
\usepackage{graphicx}
\usepackage{booktabs,multirow,makecell}
\usepackage{subcaption}
\usepackage[font=small]{caption}
\usepackage{xcolor}
\usepackage{enumitem}
\usepackage{algorithm}
\usepackage{algorithmic}
\usepackage{wrapfig}
\usepackage{url}
\usepackage[numbers,square,sort&compress]{natbib}
\usepackage[hidelinks]{hyperref}
\usepackage[capitalize,noabbrev]{cleveref}

\theoremstyle{definition}

\usepackage{amsmath,amsfonts,bm}

\def\eqref#1{equation~\ref{#1}}

\def\1{\bm{1}}

\DeclareMathAlphabet{\mathsfit}{\encodingdefault}{\sfdefault}{m}{sl}
\SetMathAlphabet{\mathsfit}{bold}{\encodingdefault}{\sfdefault}{bx}{n}

\newcommand{\xhdr}[1]{\noindent{{\bf #1.}}}
\newcommand{\our}{CrysLDM}

\makeatletter
\renewcommand{\@maketitle}{%
  \vbox{%
    \hsize\textwidth \linewidth\hsize
    \vskip 0.1in
    \hrule height 4pt \vskip 0.25in \vskip -\parskip
    \centering
    {\LARGE\bfseries \@title\par}
    \vskip 0.29in \vskip -\parskip \hrule height 1pt \vskip 0.09in
    \def\And{\end{tabular}\hfil\linebreak[0]\hfil
             \begin{tabular}[t]{c}\bfseries\rule{\z@}{24pt}\ignorespaces}%
    \def\AND{\end{tabular}\hfil\linebreak[4]\hfil
             \begin{tabular}[t]{c}\bfseries\rule{\z@}{24pt}\ignorespaces}%
    \begin{tabular}[t]{c}\bfseries\rule{\z@}{24pt}\@author\end{tabular}%
    \vskip 0.3in minus 0.1in
  }%
}

\renewenvironment{abstract}%
  {\vskip 0.075in\centerline{\large\bfseries Abstract}\vspace{0.5ex}\begin{quote}}%
  {\par\end{quote}\vskip 1ex}

\renewcommand{\section}{\@startsection{section}{1}{\z@}%
  {-2.0ex plus -0.5ex minus -0.2ex}{1.5ex plus 0.3ex minus 0.2ex}%
  {\large\bfseries\raggedright}}
\renewcommand{\subsection}{\@startsection{subsection}{2}{\z@}%
  {-1.8ex plus -0.5ex minus -0.2ex}{0.8ex plus 0.2ex}%
  {\normalsize\bfseries\raggedright}}
\renewcommand{\subsubsection}{\@startsection{subsubsection}{3}{\z@}%
  {-1.5ex plus -0.5ex minus -0.2ex}{0.5ex plus 0.2ex}%
  {\normalsize\bfseries\raggedright}}
\renewcommand{\paragraph}{\@startsection{paragraph}{4}{\z@}%
  {1.5ex plus 0.5ex minus 0.2ex}{-1em}{\normalsize\bfseries}}
\makeatother

\title{CrysLDM: Latent Diffusion Model for Crystal Material Generation}

\author{%
  Subhojyoti Khastagir\textsuperscript{1,}\thanks{Equal contribution.} \quad
  Kishalay Das\textsuperscript{1,}\footnotemark[1] \quad
  Pawan Goyal\textsuperscript{1} \quad
  Seung-Cheol Lee\textsuperscript{2} \\
  \textbf{Satadeep Bhattacharjee\textsuperscript{2} \quad
  Niloy Ganguly\textsuperscript{1}} \\[4pt]
  \textsuperscript{1}Indian Institute of Technology Kharagpur, India \\
  \textsuperscript{2}Indo Korea Science and Technology Center, Bangalore, India
}

\date{}

\begin{document}
\maketitle

\begin{abstract}
Generating new crystal materials with desirable chemical properties has long been a challenging task. Existing diffusion models operate in feature space, requiring complex diffusion architectures to model the joint distribution of atom types, coordinates, and lattice structures. This complexity increases the number of diffusion steps, leading to higher training and sampling costs.  In this work, we aim to generate novel crystal materials within a time- and resource-constrained setup, where existing models are not well-suited. To address this, we propose \our{}, a novel latent diffusion model for 3D crystal materials, which integrates a variational autoencoder and a diffusion model. The VAE encoder maps 3D crystal structures into a latent space, where the diffusion model operates. Since \our{} leverages a smooth, lower-dimensional latent space, it simplifies the generative process and accelerates both training and inference. Through extensive experiments on benchmark datasets and tasks, we show that \our{} generates stable and valid materials with quality comparable to state-of-the-art methods, while being significantly more efficient in terms of computational resources and time.
\end{abstract}

\section{Introduction}
\label{sec_intro}
Discovering novel 3D crystal materials with desired chemical properties remains a long-standing challenge in the materials design community. These materials have been essential to major advancements, including the development of batteries, solar cells, and semiconductors.~\cite{ butler2018machine,desiraju2002cryptic}. Unlike molecules, which are typically represented as graphs, crystal materials consist of a fundamental unit cell that repeats itself regularly on a 3D lattice~\cite{schutt2014represent}. Hence, the discovery of novel materials is highly challenging due to the vast search space of 3D crystal structures, which encompasses the number of constituent atoms and all their possible arrangements within 3D space. Historically, efforts to generate novel materials have relied highly on Density Functional Theory (DFT)~\cite{kohn1965self}, which is both resource-intensive and time-consuming simulations. Most recent advancements in equivariant diffusion models \cite{hoogeboom2022equivariant,bao2022equivariant}, have opened up a promising trajectory for the generation of novel three-dimensional periodic structures of crystal materials.

In this work, our aim is to generate novel crystal materials within a time- and resource-constrained setup. Given limited computational power and time budget, we focus on optimizing model capacity to maximize the generation of stable and valid materials. Existing diffusion models~\cite{xie2021crystal,luo2023towards,jiao2023crystal,zeni2023mattergen,jiao2024space,yang2023scalable} have some fundamental limitations, making them suboptimal for the problem in this time- and resource-constrained setup. Current diffusion models for generating new crystals operate in the feature space, modeling the joint distribution of atom types, fractional coordinates, and lattice structures (unit cells). However, this distribution is highly multimodal, with each component exhibiting an independent structure and distinct modality. Atomic fractional coordinates, following a wrapped normal transition distribution, are typically modeled using score matching~\cite{song2020score}. In contrast, atom types, as discrete data, are modeled using discrete diffusion (D3PM)\cite{austin2021structured}, while the lattice structure is modeled using denoising diffusion (DDPM)\cite{ho2020denoising}. As a result, a complex diffusion architecture is employed to model such a multimodal joint distribution, typically involving a higher number of diffusion steps to generate realistic crystal structures. Consequently, these models require extensive training and more inference or sampling steps, making them computationally intensive and time-consuming. 

In this work, we utilize a latent diffusion model to address the above limitations, enabling the rapid generation of stable and valid materials, within a given time and resource budget. We propose \textbf{Crys}tal \textbf{L}atent \textbf{D}iffusion \textbf{M}odel (\our{}), which consists of two modules: a Variational Auto Encoder(VAE) and a Diffusion Model(DM) that operates in a smoother latent space. The Variational Auto Encoder consists of an encoder and a decoder, where the encoder maps high-dimensional atom types and lattice structures into a lower-dimensional latent or embedding space, and the decoder reconstructs the original atom types and lattice structures from this latent representation. Additionally, a diffusion model is applied to learn and model the distribution of the latent representation of the crystal (encoded by the encoder), which is lower-dimensional and exhibits a much smoother distribution. A key challenge in developing generative models for 3D crystal structures is ensuring that the learned distribution satisfies periodic E(3) invariance, meaning it remains invariant to permutation, translation, rotation, and periodic transformations. To ensure this, we employ a periodic EGNN model as the encoder and decoder functions of the VAE, as well as the backbone denoising network in the diffusion model to guide the denoising process.

Operating in the latent space, \our{} offers unique advantages in generative modeling complexity over existing feature-domain diffusion models, making it more efficient in terms of time and resource consumption. Firstly, mapping atom types and lattice structures into a regularized latent space eliminates the need for separate diffusion processes to capture their distinct distributions. This allows the diffusion model to learn a much smoother distribution, simplifying the overall diffusion process and accelerating training. Moreover, mapping high dimensional atom type vectors into low dimensional space enables \our{} to conduct training and sampling with a lower dimensionality, which can also benefit the generative modeling complexity in terms of both time and resource consumption.

To the best of our knowledge, we are the first to leverage latent diffusion models for generating 3D crystal materials. We compare our proposed \our{} against two widely used state-of-the-art diffusion models for crystal material generation, using popular benchmark datasets. We find that \our{} generates novel materials with comparable validity and stability to other methods while being substantially faster during both training and inference. In specific, at sampling time, \our{} is \textbf{32x} and \textbf{11x} faster than CDVAE and DiffCSP, respectively, on the Perov-5 dataset, and \textbf{45x} and \textbf{6x} faster on the MP-20 dataset. These results make \our{} particularly well-suited for generating stable materials in time- and resource-constrained settings. Furthermore, we observe that the structures generated by \our{} are, on average, more stable than those produced by CDVAE and comparable in stability to those generated by DiffCSP. We provide the code base in the supplementary material.
\section{Related Work: Crystal Material Generation}
\label{sec_related}
Earlier efforts in generating novel periodic materials primarily focused on atomic composition while largely ignoring 3D structures. With advancements in generative models, researchers began using VAEs and GANs to generate 3D periodic structures. However, these models either represented materials as voxel images~\citep{court20203,hoffmann2019data,long2021constrained,noh2019inverse} or encoded structures as embedding vectors~\citep{kim2020generative,ren2020inverse,zhao2021high}, often neglecting stability and invariance to Euclidean and periodic transformations. Recent advancements in equivariant diffusion models have opened up a promising trajectory for the generation of novel three-dimensional periodic structures of crystal materials. CDVAE~\citep{xie2021crystal} was the first work that integrated a variational autoencoder (VAE) and powerful score-based decoder network, work directly with the atomic coordinates of the structures, and uses an equivariant graph neural network to ensure euclidean and periodic invariance. Subsequently, numerous studies ~\citep{luo2023towards,jiao2023crystal,zeni2023mattergen,jiao2024space,yang2023scalable} have utilized diffusion models to learn the joint distribution of atom types, coordinates, and lattice structures, enabling the generation of stable periodic structures for novel materials. 

More related works are given in Appendix \ref{appendix_related}.

\if{0}
In the past, there were limited efforts in creating novel periodic materials, with researchers concentrating on generating the atomic composition of periodic materials while largely neglecting the 3D structure. With the advancement of generative models, the majority of the research focuses on using popular generative models like VAEs or GANs to generate 3D periodic structures of materials, however, they either represent materials as three-dimensional voxel images~\citep{court20203,hoffmann2019data,long2021constrained,noh2019inverse}  and generate images to depict material structures (atom types, coordinates, and lattices), or they directly encode material structures as embedding vectors~\citep{kim2020generative,ren2020inverse,zhao2021high}. However, these models neither incorporate stability in the generated structure nor are invariant to any Euclidean and periodic transformations. Recent advancements in equivariant diffusion models have opened up a promising trajectory for the generation of novel three-dimensional periodic structures of crystal materials. CDVAE~\citep{xie2021crystal} was the first work that integrated a variational autoencoder (VAE) and powerful score-based decoder network, work directly with the atomic coordinates of the structures and uses an equivariant graph neural network to ensure euclidean and periodic invariance. Subsequently, numerous studies ~\citep{luo2023towards,jiao2023crystal,zeni2023mattergen,jiao2024space,yang2023scalable} have utilized diffusion models to learn the joint distribution of atom types, coordinates, and lattice structures, enabling the generation of stable periodic structures for novel materials.
\fi

\section{Background}
\label{sec_prelim}
\subsection{Crystal Structure Representation}
\label{prelim_crystal}
The 3D structure of a crystal material can be modeled by a minimal \textit{Unit Cell}, which gets repeated infinite times in three-dimensional space on a regular lattice to form the periodic crystal structure. Given a material with $N$ number of atoms in its unit cell, we can describe the unit cell by two matrices: Atom Type matrix $\textbf{\textit{A}}=[\textbf{\textit{a}}_1, \textbf{\textit{a}}_2, \ldots, \textbf{\textit{a}}_N]^T\in\mathbb{R}^{N\times k}$, which denotes a set of atomic type in one hot representation (k denotes maximum number of possible atom types), and Coordinate Matrix $\textbf{\textit{X}}=[\textbf{\textit{x}}_1,\textbf{\textit{x}}_2,...,\textbf{\textit{x}}_N]^T \in \mathbb{R}^{N \times 3}$ denotes fractional coordinate positions of atoms, where $\textbf{\textit{x}}_i \in \mathbb{R}^{3}$ corresponds to coordinates of $i^{th}$ atom in the unit cell. Further, there is an additional \textit{Lattice Matrix} $\textbf{\textit{L}}=[\textbf{\textit{l}}_1,\textbf{\textit{l}}_2,\textbf{\textit{l}}_3]^T \in \mathbb{R}^{3 \times 3}$, which describes how a unit cell repeats itself in the 3D space towards $\textbf{\textit{l}}_1,\textbf{\textit{l}}_2$ and $\textbf{\textit{l}}_3$ direction to form the periodic 3D structure of the material. Formally, a given material can be defined as $ \textbf{\textit{M}}=(\textbf{\textit{A}},\textbf{\textit{X}},\textbf{\textit{L}})$ and we can represent its infinite periodic structure as $\mathbf{\hat{X}}  = \{ \hat{x}_i |  \hat{x}_i = \textit{x}_i + \sum_{j=1}^{3} k_j{l}_j \}$; $\mathbf{\hat{A}}  = \{ \hat{a}_i |  \hat{a}_i = {a}_i \}$ where $k_1,k_2,k_3, i \in Z, 1 \leq i \leq N$.
\subsection{Invariances in Crystal Structure}
\label{prelim_inv}
The basic idea of using generative models for crystal generation is to learn the underlying data distribution of material structure $f(\textbf{\textit{M}})$. Since crystal materials satisfy physical symmetry properties \cite{dresselhaus2007group,zee2016group}, one of the major challenges here is the learned distribution must satisfy periodic E(3) invariance i.e. invariance to permutation, translation, rotation, and periodic transformations. 1) \textbf{\textit{Permutation Invariance :}} If we permute the indices of constituent atoms it will not change the material. 2) \textbf{\textit{Translation Invariance :}} If we translate the atom coordinates by a random vector it will not change the structure of the material. 3) \textbf{\textit{Rotational Invariance :}} If we rotate the atom coordinates and lattice matrix, the material remains unchanged. 4) \textbf{\textit{Periodic Invariance :}} Finally, since the atoms in the unit cell can periodically repeat itself infinite times along the lattice vector, there can be many choices of unit cells and coordinate matrices representing the same material. (More details in Appendix \ref{appendix_prelim_inv})
\subsection{Problem Formulation}
\label{prelim_problem}
In this work, we consider generative modeling of 3D crystal geometries from scratch, to discover new stable materials. Formally, given a dataset $\mathcal{M}=\{ \textbf{\textit{M}}_i \}$, containing crystal structure {$\textbf{\textit{M}}_i = (\textbf{\textit{A}}_i,\textbf{\textit{X}}_i,\textbf{\textit{L}}_i)$}, the goal is to capture the underlying data distribution $f(\textbf{\textit{M}})$ via learning a generative model $p_{\theta} (\textbf{\textit{M}})$, where $\theta$ is a set of learnable parameters. While training, we need $p_{\theta}$ to ensure that the learned distribution is invariant to different symmetry transformations mentioned in Section \ref{prelim_inv}. Once trained, the learned generative model can sample a valid and stable structure of the material, that is invariant to different symmetry transformations.
\subsection{Diffusion Models for Crystal Materials}
\label{prelim_diffusion_crystal}
Diffusion models are popular generative models that are formulated using a T steps Markov Chain. Given a data point $\mathbf{d}_0$, the forward diffusion process gradually corrupts the data point over $T$ steps, by adding a small amount of gaussian noise at each step:
\begin{equation}
    \label{eq:forward}
        q(\mathbf{d}_{t}|\mathbf{d}_{0}) = \mathcal{N} (\mathbf{d}_{t}|\sqrt{\bar{\alpha}_{t}}\mathbf{d}_{0}, \ (1 \ - \ \bar{\alpha}_{t})\mathbf{I} )
\end{equation}
where, $\bar{\alpha}_{t} = \prod_{k=1}^{t} \alpha_{k}$, $\alpha_{t} = 1-\beta_{t}$ and $\{ \beta_{t} \in (0,1) \}^{T}_{t=1}$ controls the variance of diffusion step following certain noise scheduler. Further, the reverse denoising process, which is parameterized, begins with a gaussian noise input $\mathbf{d}_T \sim \mathcal{N}(\mathbf{0},\mathbf{I})$ and incrementally denoises the intermediate noisy variables $\mathbf{d}_{T:1}$ to approximate the clean data $\mathbf{d}_0$ following the target data distribution:
\begin{equation}
    \label{eq:backward}
    \begin{split}
        p_\theta(\mathbf{d}_{t-1} | \mathbf{d}_{t}) = \mathcal{N} \big \{ \mathbf{d}_{t-1} \ ; \ \mu_\theta (\mathbf{d}_{t-1},t), \rho^2_t \mathbf{I} \big \} 
    \end{split}
\end{equation}
where $\rho_t$ is a predefined variance and mean $\mu_\theta$ is typically modeled using some neural network (U-Net for images). However, leveraging diffusion models to generate new crystal materials is challenging due to the highly multi-modal nature of their joint distribution, where each component has an independent structure and a distinct modality. On one hand, atom types are discrete, while lattice parameters are continuous; additionally, atomic fractional coordinates are continuous but exhibit periodicity. As a result, each variable necessitates a independent diffusion framework to accommodate its unique structure.

\xhdr{Diffusion on Atom Type (\textbf{\textit{A}})} Atom Type Matrix $\textbf{\textit{A}} \in \mathbb{R}^{N \times k}$ can be considered as N discrete variables belonging to k classes and discrete diffusion model (D3PM)~\citep{austin2021structured} can be leveraged for diffusion on \textbf{\textit{A}}. In specific, with \textbf{\textit{a}} as the one-hot representation of atom \textit{a}, the transition probability for the forward process is $q(\textbf{\textit{a}}_t | \textbf{\textit{a}}_{t-1}) = Cat(\textbf{\textit{a}}_t;\textbf{\textit{p}} = \textbf{\textit{a}}_{t-1}\textbf{\textit{Q}}_{t})$, where $Cat(\textbf{\textit{a}};\textbf{\textit{p}})$ is a categorical distribution over \textbf{\textit{a}} with probabilities \textbf{\textit{p}} and $\textbf{\textit{Q}}_{t}$ is the Markov transition matrix at time step t, defined as $[\textbf{\textit{Q}}_{t}]_{i,j} = q(\textbf{\textit{a}}_t = i | \textbf{\textit{a}}_{t-1} = j)$. Different choices of $\textbf{\textit{Q}}_{t}$ and corresponding stationary distributions are proposed by~\citep{austin2021structured} which provides flexibility to control the data corruption and denoising process.
\begin{figure*}
	\centering
	\includegraphics[width=0.8\columnwidth]{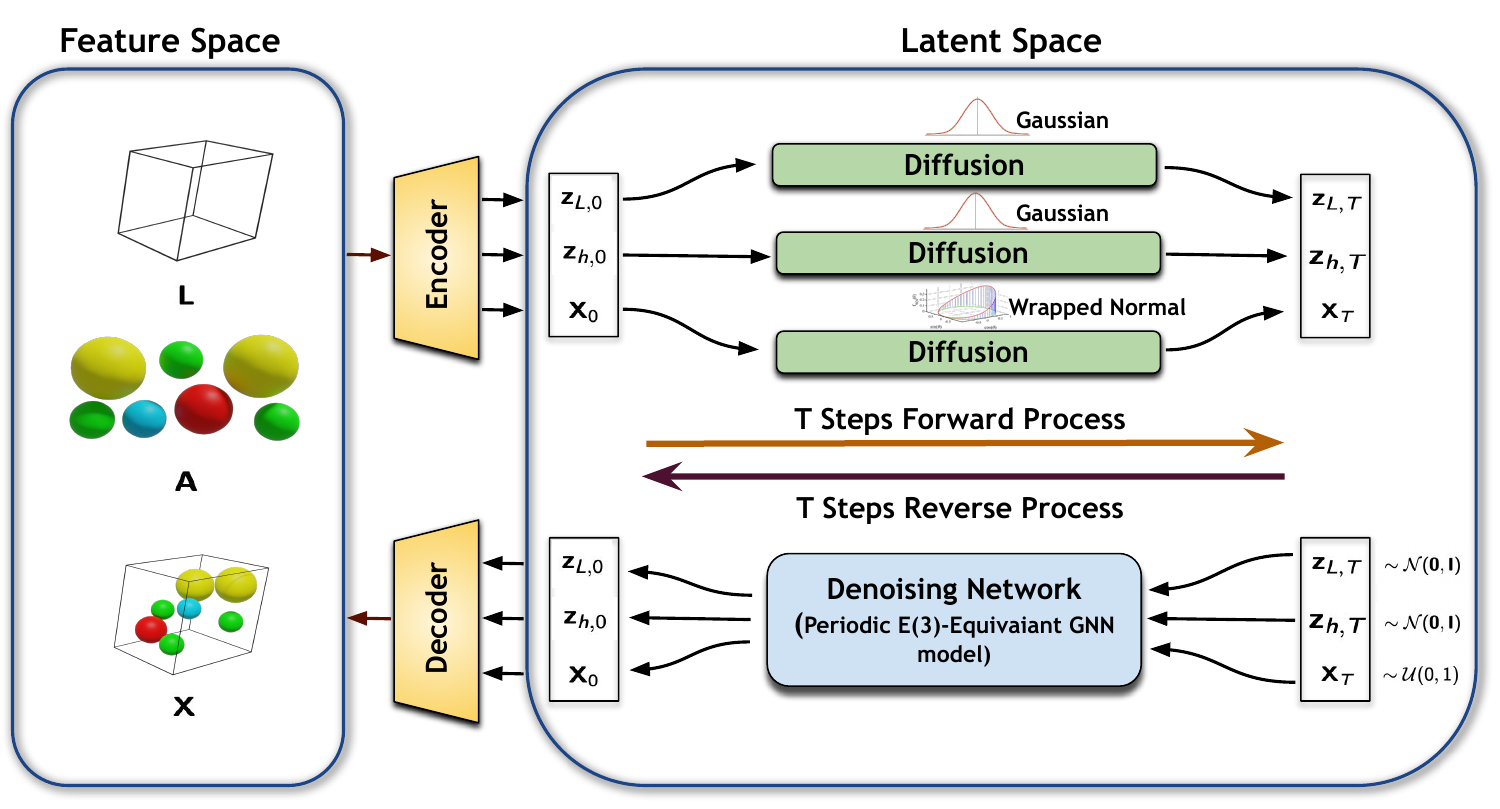}
	\caption{Model Architecture of our proposed \our{}.}
	\label{fig:architecture}
\end{figure*}

\xhdr{Diffusion on Atom Coordinates (\textbf{\textit{X}})} Coordinate Matrix $\textbf{\textit{X}}=[\textbf{\textit{x}}_1,\textbf{\textit{x}}_2,...,\textbf{\textit{x}}_N]^T \in \mathbb{R}^{N \times 3}$ contains fractional coordinates of constituent atoms, that resides in quotient space $\mathbb{R}^{N \times 3} / \mathbb{Z}^{N \times 3}$ induced by the crystal periodicity. Hence, it is not suitable to apply DDPM to model \textbf{\textit{X}}, since the gaussian distribution used in DDPM is unable to model the cyclical and bounded domain of \textbf{\textit{X}}. Hence at each step of forward diffusion, noise sampled from Wrapped Normal (WN) distribution~\citep{de2022riemannian} is added to \textbf{\textit{X}} and during denoising  Score Matching Network~\citep{song2019generative,song2020improved} is leveraged to model underlying transition probability.

\xhdr{Diffusion on Lattice (\textbf{\textit{L}})} Lattice Matrix $\textbf{\textit{L}}=[{l}_1,{l}_2,{l}_3]^T \in \mathbb{R}^{3 \times 3}$ is a global feature of the material which determines the shape and symmetry of the unit cell structure. Since \textbf{\textit{L}} is in continuous space, we leverage the idea of the Denoising Diffusion Probabilistic Model (DDPM)~\citep{ho2020denoising} for diffusion on \textbf{\textit{L}}.

During the reverse denoising process, a key challenge is ensuring that the learned distribution of material structures adheres to periodic E(3) invariance. To address this, existing works have utilized variants of periodic-E(3)-equivariant GNN models, such as GemNet~\citep{gasteiger2021gemnet}, DimeNet~\citep{gasteiger2020directional}, or  CSPNet~\citep{jiao2023crystal}, as backbone denoising networks to guide the denoising process. Training the denoising network involves an aggregated objective function that combines cross-entropy loss, score matching loss, and $l_2$ for atom types, coordinates, and lattice parameters, respectively:
Due to the complexity of the diffusion architecture required to model such a multimodal joint distribution, a higher number of diffusion steps is typically needed to generate realistic crystal structures. As a result, these models demand substantial training effort and involve more inference or sampling steps, making them both computationally intensive and time-consuming.
\section{Methodology}
\label{sec_method}
In this section, we provide a detailed overview of our proposed methodology, \textbf{Crys}tal \textbf{L}atent \textbf{D}iffusion \textbf{M}odel (\our{}). The framework comprises two main components: a Variational Autoencoder (VAE) and a Diffusion Model (DM) that operates within a smoother latent space. We will first describe the detailed architecture of both modules, followed by their respective training and sampling processes.
\subsection{Variational Autoencoder (VAE)}
\label{sec_ae}
Our first objective is to encode 3D crystal geometry into a lower-dimensional latent space that is meaningful and preserves all the physical symmetries of the crystal structure. To achieve this, we utilize a variational autoencoder framework comprising a GNN encoder, \(\mathcal{E}_\phi\), and a GNN decoder, \(\mathcal{D}_\psi\). The encoder takes the crystal material \(\textbf{\textit{M}} = (\textbf{\textit{A}}, \textbf{\textit{X}}, \textbf{\textit{L}})\) as input and encodes the atomic types and lattice structure into the latent space:
\begin{equation}
    \label{eq:enc}
       \textbf{\textit{z}}_h, \textbf{\textit{z}}_L  = \mathcal{E}_\phi(\textbf{\textit{A}}, \textbf{\textit{X}}, \textbf{\textit{L}})
\end{equation}
where $\textbf{\textit{z}}_h$ and $\textbf{\textit{z}}_L$ are latent representation of constituent nodes(atoms) and lattice structure respectively. Note, we did not encode atomic fractional coordinates $\textbf{\textit{X}}$ into the latent space primarily for two main reasons: first, atomic coordinates are inherently low-dimensional, and second, empirical observations showed that encoding atomic coordinates into the latent space diminishes their physical significance, adversely affecting the message-passing process of GNNs and subsequently degrading the model's performance. Further, the decoder \(\mathcal{D}_\psi\) is trained to reconstruct the atomic types and lattice structure from the latent representations:
\begin{equation}
    \label{eq:dec}
       \tilde{\textbf{\textit{A}}}, \tilde{\textbf{\textit{L}}}  = 
       \mathcal{D}_\psi(\textbf{\textit{z}}_h, \textbf{\textit{X}}, \textbf{\textit{z}}_L)
\end{equation}
A key criterion in designing this variational autoencoder is that the learned latent space must preserve the physical symmetries of the crystal structure and satisfy periodic E(3) invariance. To achieve this, we propose incorporating equivariance into the autoencoder design by using 3D equivariant graph neural networks (EGNNs) to implement both the encoder \(\mathcal{E}_\phi\) and the decoder \(\mathcal{D}_\psi\). The whole auto-encoder network is trained end to end using a regularized reconstruction loss:
\begin{equation}
    \begin{split}
    \label{eq:ae_loss}
        \mathcal{L}_{VAE} = \mathcal{L}^{\textbf{\textit{A}}}_{recon} + \mathcal{L}^{\textbf{\textit{L}}}_{recon} + \lambda \mathcal{L}_{reg} \\
        \mathcal{L}_{reg} = {d}_{KL} \{ q_{\phi}( \textbf{\textit{z}}_h, \textbf{\textit{z}}_L | \textbf{\textit{A}}, \textbf{\textit{X}}, \textbf{\textit{L}}) \: || \: p(\textbf{\textit{z}}_h, \textbf{\textit{z}}_L) \}
    \end{split}
\end{equation}
Here, \(\mathcal{L}^{\textbf{\textit{A}}}_{recon}\) and \(\mathcal{L}^{\textbf{\textit{L}}}_{recon}\) represent the reconstruction losses for atom types and lattice structure, respectively. By design, we use cross-entropy loss for \(\textbf{\textit{A}}\) and \(l_2\) loss for \(\textbf{\textit{L}}\). $\mathcal{L}_{reg}$ denotes KL divergence (Kullback–Leibler divergence) that measures how much the learned latent distribution $q_{\phi}( \textbf{\textit{z}}_h, \textbf{\textit{z}}_L | \textbf{\textit{A}}, \textbf{\textit{X}}, \textbf{\textit{L}})$ differs from the prior distribution $p(\textbf{\textit{z}}_h, \textbf{\textit{z}}_L)$ (commonly a standard gaussian distribution). This regularization term constrains the variance of latent embeddings, making them more stable and suitable for learning latent diffusion models (LDMs).
\subsection{Latent Diffusion Model(LDM)}
\label{sec_ldm}
The encoder function \(\mathcal{E}_\phi\) of the crystal autoencoder allows us to encode crystal materials into a smoother and lower-dimensional latent space. Building on this, we leverage a latent diffusion model to capture the distribution of crystal latent or embedding space. Given latent representation of a crystal material $ $ as $\textbf{\textit{M}}_0 = (\textbf{\textit{z}}_{h,0}, \textbf{\textit{X}}_0, \textbf{\textit{z}}_{L,0})$, we define a forward diffusion process through a Markov chain over T steps to diffuse  $\textbf{\textit{z}}_h$, \textbf{\textit{X}}, $\textbf{\textit{z}}_L$ independently as follows :
\begin{equation}
    \label{eq:joint_dist}
    \begin{split}
        q(\textbf{\textit{z}}_{h,t},\textbf{\textit{X}}_{t},\textbf{\textit{z}}_{L,t} \ | \ 
        \textbf{\textit{z}}_{h,t-1},\textbf{\textit{X}}_{t-1},\textbf{\textit{z}}_{L,t-1}) = \\
        q(\textbf{\textit{z}}_{h,t} \ | \ \textbf{\textit{z}}_{h,t-1}) 
        q(\textbf{\textit{X}}_{t} \ | \ \textbf{\textit{X}}_{t-1}) 
        q(\textbf{\textit{z}}_{L,t} \ | \ \textbf{\textit{z}}_{L,t-1}) 
        \: \ t=1,2,...T
    \end{split}
\end{equation}

\xhdr{Diffusion on $\textbf{\textit{z}}_h$ and $\textbf{\textit{z}}_L$} Since both atom types and lattice structure are projected into smooth latent space using The encoder function \(\mathcal{E}_\phi\), we don't need separate sophisticated diffusion process to model them. Rather, as both $\textbf{\textit{z}}_h$ and $\textbf{\textit{z}}_L$ are in continuous space, we can use Denoising Diffusion Probabilistic Model (DDPM)~\citep{ho2020denoising} to model them. Given $\textbf{\textit{z}}_{h,0} \sim p(\textbf{\textit{z}}_h)$ and  $\textbf{\textit{z}}_{L,0} \sim p(\textbf{\textit{z}}_L)$, the forward diffusion process iteratively diffuses both over T timesteps
through transition probabilities $q( \textbf{\textit{z}}_{h,t} |  \textbf{\textit{z}}_{h,0} )$ and $q( \textbf{\textit{z}}_{L,t} |  \textbf{\textit{z}}_{L,0} )$ respectively. At each $t^{th}$ step, we can derive these probabilities as follows :
\begin{equation}
    \label{eq:h_l_forward}
    \begin{split}
        q(\textbf{\textit{z}}_{h,t} \ | \ \textbf{\textit{z}}_{h,0}) = \mathcal{N} (\textbf{\textit{z}}_{h,t} \ | \ \sqrt{\bar{\alpha}_{t}}\textbf{\textit{z}}_{h,0}, \ (1 \ - \ \bar{\alpha}_{t})\mathbf{I} ) \\
        q(\textbf{\textit{z}}_{L,t} \ | \ \textbf{\textit{z}}_{L,0}) = \mathcal{N} (\textbf{\textit{z}}_{L,t} \ | \ \sqrt{\bar{\alpha}_{t}}\textbf{\textit{z}}_{L,0}, \ (1 \ - \ \bar{\alpha}_{t})\mathbf{I})
    \end{split}
\end{equation}
where, $\bar{\alpha}_{t} = \prod_{k=1}^{t} \alpha_{k}$, $\alpha_{t} = 1-\beta_{t}$ and $\{ \beta_{t} \in (0,1) \}^{T}_{t=1}$ controls the variance of diffusion step following certain variance scheduler. By reparameterization, we can rewrite equation \ref{eq:h_l_forward} as:
\begin{equation}
    \label{eq:h_l_forward_reparam}
    \begin{split}
        \textbf{\textit{z}}_{h,t} = \sqrt{\bar{\alpha}_{t}}\textbf{\textit{z}}_{h,0} +  \sqrt{1-\bar{\alpha}_{t}}\bm{\epsilon}_{\textbf{\textit{z}}_{h}} \\
        \textbf{\textit{z}}_{L,t} = \sqrt{\bar{\alpha}_{t}}\textbf{\textit{z}}_{L,0} +  \sqrt{1-\bar{\alpha}_{t}}\bm{\epsilon}_{\textbf{\textit{z}}_{L}}
    \end{split}
\end{equation}
where, $\bm{\epsilon}_{\textbf{\textit{z}}_{h}}, \bm{\epsilon}_{\textbf{\textit{z}}_{L}} \sim \mathcal{N}(\mathbf{0},\mathbf{I})$ are noises.  After T such diffusion steps, noisy  $\textbf{\textit{z}}_{h,T},\textbf{\textit{z}}_{L,T} $ is generated, which follows prior noise distribution $\sim \mathcal{N}(\mathbf{0},\mathbf{I})$.

In the reverse denoising process, given noisy $\textbf{\textit{z}}_{h,T},\textbf{\textit{z}}_{L,T} \sim \mathcal{N}(\mathbf{0},\mathbf{I})$ we reconstruct true latent representation of atom and lattice $\textbf{\textit{z}}_{h,0},\textbf{\textit{z}}_{L,0}$ thorough iterative denoising step via learning reverse conditional distribution, which we formulate as follows :
\begin{equation}
    \label{eq:h_l_backward}
    \begin{split}
        p(\textbf{\textit{z}}_{h,t-1} | \textbf{\textit{M}}_{t}) = 
        \mathcal{N} \big \{ \textbf{\textit{z}}_{h,t-1} \ | \ \mu^{\textbf{\textit{z}}_{h}}(\textbf{\textit{M}}_{t}),\rho^2_t\mathbf{I} \big \} \\
        p(\textbf{\textit{z}}_{L,t-1} | \textbf{\textit{M}}_{t}) = 
        \mathcal{N} \big \{ \textbf{\textit{z}}_{L,t-1} \ | \ \mu^{\textbf{\textit{z}}_{L}}(\textbf{\textit{M}}_{t}),\rho^2_t\mathbf{I} \big \}
    \end{split}
\end{equation}
where $\mu^{\textbf{\textit{z}}_{h}}(\textbf{\textit{M}}_{t}) = \frac{1}{\sqrt{\alpha_{t}}}\big(\textbf{\textit{z}}_{h,t} - \frac{1 - \alpha_{t}}{\sqrt{1 - \bar{\alpha}_{t}}} \ \hat{\bm{\epsilon}}_{\textbf{\textit{z}}_{h}} (\textbf{\textit{M}}_{t},t)\big)$, $\mu^{\textbf{\textit{z}}_{L}}(\textbf{\textit{M}}_{t}) = \frac{1}{\sqrt{\alpha_{t}}}\big(\textbf{\textit{z}}_{L,t} - \frac{1 - \alpha_{t}}{\sqrt{1 - \bar{\alpha}_{t}}} \ \hat{\bm{\epsilon}}_{\textbf{\textit{z}}_{L}} (\textbf{\textit{M}}_{t},t)\big)$ and  $\textbf{\textit{M}}_{t} = (\textbf{\textit{z}}_{h,t},\textbf{\textit{X}}_{t},\textbf{\textit{z}}_{L,t})$.  Intuitively, $\hat{\bm{\epsilon}}_{\textbf{\textit{z}}_{h}}, \hat{\bm{\epsilon}}_{\textbf{\textit{z}}_{L}}$ are the denoising terms that needs to be subtracted from $\textbf{\textit{z}}_{h,t}$ and $\textbf{\textit{z}}_{L,t}$ to generate $\textbf{\textit{z}}_{h,t-1}$ and $\textbf{\textit{z}}_{L,t-1}$ respectively. We use a denoising network $f_\theta(\textbf{\textit{M}}_{t},t)$ to model these noise terms. Following the simplified training objective proposed by ~\cite{ho2020denoising}, we train the denoising network using following $l_2$ losses :
\begin{equation}
    \label{eq:lattice_loss}
    \begin{split}
        \mathcal{L}_{type} = \mathbb{E}_{\bm{\epsilon}_{\textbf{\textit{z}}_{h}},t \sim \mathcal{U}(1,T)}
        \lVert  \bm{\epsilon}_{\textbf{\textit{z}}_{h}} \ - \ \hat{\bm{\epsilon}}_{\textbf{\textit{z}}_{h}} {\rVert}^2_2 \\
        \mathcal{L}_{lattice} = \mathbb{E}_{\bm{\epsilon}_{\textbf{\textit{z}}_{L}},t \sim \mathcal{U}(1,T)}
        \lVert  \bm{\epsilon}_{\textbf{\textit{z}}_{L}} \ - \ \hat{\bm{\epsilon}}_{\textbf{\textit{z}}_{L}} {\rVert}^2_2 
    \end{split}
\end{equation}

\xhdr{Diffusion on $\textbf{\textit{X}}$} As discussed earlier in Section \ref{sec_ae}, atomic fractional coordinates are not being projected into the latent space. Therefore, we will utilize the conventional feature space diffusion process to model $\textbf{\textit{X}}$. Atomic fractional coordinates in crystal material live in quotient space $\mathbb{R}^{N \times 3} / \mathbb{Z}^{N \times 3}$ induced by the crystal periodicity. At each step of forward diffusion, we add noise sample from Wrapped Normal (WN) distribution ~\cite{de2022riemannian} to \textbf{\textit{X}} and during backward diffusion leverage Score Matching Diffusion Networks ~\cite{song2019generative,song2020improved} to model underlying transition probability:
\begin{equation}
    \label{eq:type_loss}
    \begin{split}
        q(\textbf{\textit{X}}_{t} \ | \ \textbf{\textit{X}}_{0}) = \mathcal{N}_W (\textbf{\textit{X}}_{t} \ | \ \textbf{\textit{X}}_{0}, \sigma_t^2 \mathbf{I})
    \end{split}
\end{equation}
In specific, at each $t^{th}$ step of diffusion, we derive $\textbf{\textit{X}}_{t}$ as : 
$\textbf{\textit{X}}_{t} = \Gamma_w(\textbf{\textit{X}}_{0} + \bm{\sigma_t}\bm{\epsilon}_{\textbf{\textit{X}}})$
where, $\bm{\epsilon}_{\textbf{\textit{X}}}$ is a noise, sampled from $\mathcal{N}(\mathbf{0},\mathbf{I})$, $\bm{\sigma_t}$ is the noise scale following exponential scheduler and $\Gamma_w(.)$ is a truncation function. Given a fractional coordinate matrix X, truncation function $\Gamma_w(\textbf{\textit{X}}) = (\textbf{\textit{X}} - \lfloor \textbf{\textit{X}} \rfloor)$ returns the fractional part of each element of \textbf{\textit{X}}. As argued in ~\cite{jiao2023crystal}, $q(X_t|X_0)$ is periodic translation equivariant, and approaches uniform distribution $\mathcal{U}(0,1)$ for sufficiently large values of $\sigma_T$. Hence during the backward denoising process, we first sample $\textbf{\textit{X}}_{T} \sim \mathcal{U}(0,1)$ and iteratively denoise via score network for T steps to recover back the true fractional coordinates $\textbf{\textit{X}}_{0}$. We use the denoising network $f_\theta(\textbf{\textit{M}}_{t},t)$ to model the backward diffusion process, which is trained using the following score-matching objective function :
\begin{equation}
    \label{eq:appendix_coord_loss}
        \mathcal{L}_{coord} = \mathbb{E}_{\substack{\textbf{\textit{X}}_{t} \sim q(\textbf{\textit{X}}_{t} | \textbf{\textit{X}}_{0}) \\
        t \sim \mathcal{U}(1,T)}}
        \lVert  \nabla_{\textbf{\textit{X}}_{t}} \text{log} q(\textbf{\textit{X}}_{t} | \textbf{\textit{X}}_{0}) - \hat{\bm{\epsilon}}_{\textbf{\textit{X}}}(\textbf{\textit{M}}_{t},t) {\rVert}^2_2
\end{equation}
where $\nabla_{\textbf{\textit{X}}_{t}} \text{log} q(\textbf{\textit{X}}_{t} | \textbf{\textit{X}}_{0}) \propto \sum_{\textbf{\textit{K}} \in \mathbb{Z}^{N \times 3}} \text{exp}(- \ \frac{\lVert \textbf{\textit{X}}_{t} - \textbf{\textit{X}}_{0} + \textbf{\textit{K}} {\rVert}^2_F }{2\bm{\sigma_t}^2})$ is the score function of transitional distribution and $\hat{\bm{\epsilon}}_{\textbf{\textit{X}}}(\textbf{\textit{M}}_{t},t)$ denoising term.

\xhdr{Denoising Network} The goal of the denoising network during the reverse denoising process is to iteratively remove noise from the noisy crystal representation (sampled from gaussian noise), transforming it into a realistic crystal latent representation. A crucial requirement for this process is that the learned distribution of material structures must adhere to periodic E(3) invariance. As established in the literature \cite{xu2022geodiff}, 
the learned distribution $p(\textbf{\textit{M}}_0)$ of the denoising model will satisfy periodic E(3) invariance, provided the prior distribution $p(\textbf{\textit{M}}_T)$ is invariant and the neural network parameterizing the transition probability $q(\textbf{\textit{M}}_{t-1} | \textbf{\textit{M}}_t)$ is equivariant to permutation, translation, rotation, and periodic transformations. To fulfill this requirement, we employed 3D equivariant graph neural networks (EGNNs) to implement the denoising network. In practice, we extend the CSPNet architecture~\cite{jiao2023crystal}, which was originally designed for the crystal structure prediction (CSP) task. CSPNet is based on the EGNN and satisfies the periodic E(3) invariance condition for periodic crystal structures(Details in Appendix \ref{app_cspnet}). The denoising network is trained using a combined loss: $ \mathcal{L}_{LDM} = \lambda_{L} \mathcal{L}_{lattice} + \lambda_{A} \mathcal{L}_{type} + \lambda_{X} \mathcal{L}_{coord}$
where, the hyperparameters $\lambda_{L}$, $\lambda_{A}$, and $\lambda_{X}$ control the relative weighting of these loss components.
\newcommand\Tstrut{\rule{0pt}{2.6ex}}
\newcommand\Bstrut{\rule[-0.9ex]{0pt}{0pt}}

\begin{table*}
\centering
\resizebox{1.0\textwidth}{!}{
\begin{tabular}{|c|c|c|cc|cc|cc|}
  \hline
  \multirow{2}{*}{Dataset} & \multirow{2}{*}{Method} & \multirow{2}{*}{Steps} & \multicolumn{2}{c|}{Validity(\%) $\uparrow$} & \multicolumn{2}{c|}{Property $\downarrow$} & Stability $\uparrow$ & Cost $\downarrow$ \Tstrut\Bstrut\\
  \cline{4-7}
   & & & Compositional & Structural & \# Elements & Density & Rate (\%) & Steps / Stable \Tstrut\Bstrut\\
  \hline
  \multirow{3}{*}{Perov} & CDVAE & 5000 & 98.59 & \textbf{100} & 0.0264 & 0.1258 & \textbf{2.5} & 200 \Tstrut\Bstrut\\
  \cline{2-9}
   & DiffCSP & 1000 & \textbf{98.85} & \textbf{100} & \textbf{0.0263} & \textbf{0.111} & 1.5 & 66.67 \Tstrut\Bstrut\\
  \cline{2-9}
   & \our{} & 100 & 97.81 & \textbf{100} & 0.045 & 1.21 & 1.6 & \textbf{6.25} \Tstrut\Bstrut\\
  \hline
  \multirow{3}{*}{MP-20} & CDVAE & 5000 & \textbf{86.7} & \textbf{100} & 0.2778 & 0.6875 & 33.2 & 15.06 \Tstrut\Bstrut\\
  \cline{2-9}
   & DiffCSP & 1000 & 83.25 & \textbf{100} & \textbf{0.1247} & 0.3502 & \textbf{47.6} & 2.1 \Tstrut\Bstrut\\
  \cline{2-9}
   & \our{} & 100 & 83.02 & 99.87 & 0.5149 & \textbf{0.3115} & 30.3 & \textbf{0.33} \Tstrut\Bstrut\\
  \hline
\end{tabular}
}
\caption{Summary of results on Ab Initio Crystal Generation task of \our{} and different baseline models on Perov-5 and MP-20 dataset. The best performances are highlighted in bold.}
\label{tbl-gen}
\end{table*}

\xhdr{Training and Sampling}
Next, we outline the training and sampling process of \our{}. Following previous works on latent diffusion models in other domains ~\citep{sinha2021d2c, rombach2022high, xu2023geometric}, we have adopted a two-stage training strategy. We first train the VAE with a regularized reconstruction loss (\ref{eq:ae_loss}), followed by training the Latent Diffusion Model using the combined diffusion loss $\mathcal{L}_{LDM}$.\\
During sampling, our setup first requires determining the number of constituent atoms (\(N\)) in the material. Following common practice in the literature~\citep{jiao2023crystal}, we begin by estimating the distribution of atom counts, \( p(N) \), across different materials in the training set. We then sample \( N \sim p(N) \) and generate atom latent features and coordinates of size \( N \). Next, we generate atomic fractional coordinates along with latent embeddings of atoms and lattices using the latent diffusion models. These embeddings are then fed into the decoder $\mathcal{D}_\psi$ of the VAE to generate 3D structure of a realistic material. The training and sampling algorithms are given in Appendix \ref{app_algo}.

\xhdr{Advantage of \our{}}
By design, operating a diffusion model in the latent space offers several inherent advantages. First, since we utilize a variational autoencoder (VAE) trained with a regularized reconstruction loss, the latent space becomes more compact and smooth, enhancing the training efficiency of the diffusion model. Second, in feature space, atom types and lattice structures belong to different modalities, requiring separate diffusion processes to accommodate their distinct distributions. However, in \our{}, both are encoded into a unified smooth latent space, simplifying the overall diffusion process and accelerating training. Finally, mapping high-dimensional atom-type vectors into a lower-dimensional space allows \our{} to perform training and sampling with reduced dimensionality. This not only improves generative modeling efficiency but also reduces computational costs in terms of both time and resource consumption.
\section{Experiments}
\xhdr{Evaluation Metric} Following prior works, we evaluate the performance of \our{} and baseline models in generating novel material structures using a diverse set of metrics, categorized under \textbf{Validity}, \textbf{Property Statistics}, and \textbf{Stability} measure. Under \textbf{Validity}, in line with previous studies ~\cite{court20203, xie2021crystal}, we assess both structural and compositional validity. Structural validity represents the percentage of generated crystals with valid periodic structures, while compositional validity refers to the percentage of structures with correct atom types. A structure is considered valid if the shortest distance between any pair of atoms exceeds 0.5 Å, and its composition is deemed valid if the overall charge remains neutral, as determined by SMACT~\cite{davies2019smact}. Additionally, we evaluate the similarity between the generated materials and those in the test set using various \textbf{Property Statistics}, where we compute the earth mover’s distance (EMD) between the distributions in element number (\# Elem) and density ($\rho$, unit g/cm3). Finally, we evaluate \textbf{Stability} of our generated materials, which is based on energy above hull ($E^{hull}$) calculations. To evaluate the stability, we first sample 1000 materials and use M3GNet~\citep{Chen2022} to relax the structures of the generated materials and approximate force, energy, and stress within the crystal unit. We then classify final relaxed structures with a predicted energy above hull of less than 0.1 eV/atom as stable materials. We report \textbf{Stability Rate} as \% of stable materials out of 1000 samples.

\begin{figure}
    \centering
    \includegraphics[width=0.48\columnwidth]{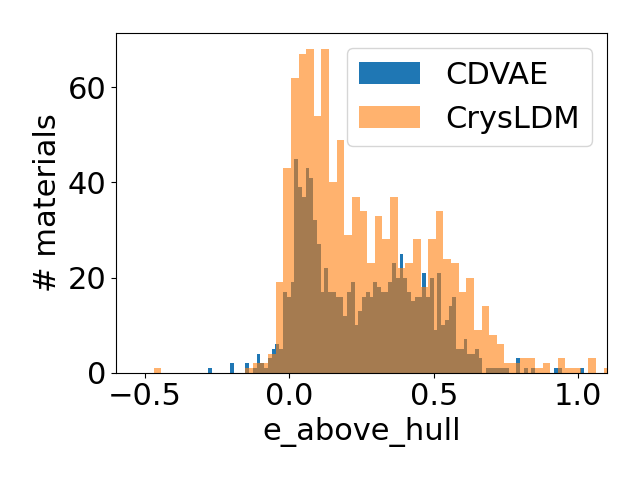}
    \includegraphics[width=0.48\columnwidth]{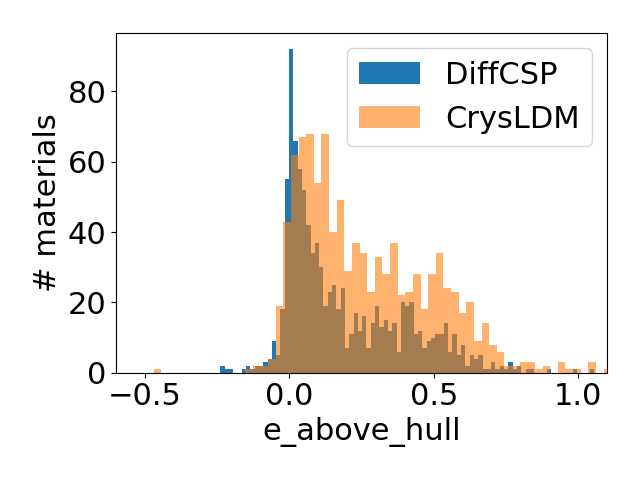}
    \caption{Histogram of \(E^{hull}\) distribution for relaxed structures generated by different models.}
    \label{fig:hist}
\end{figure}

\xhdr{Results}
We present the material generation results for both datasets in Table (\ref{tbl-gen}). We observe that across all metrics, \our{} demonstrates competitive performance compared to baseline models, highlighting the strong generative capabilities of our proposed latent diffusion framework. Furthermore, given the time- and resource-constrained setup, our goal is to minimize the overall computational cost of material generation. To assess this, we report the \textbf{Stability Cost}, defined as the number of integration steps per stability rate (\# Integration Steps /\# Stable Materials), which quantifies the average number of sampling steps required to generate a stable material. We observe that, compared to baseline models, \our{} requires fewer sampling steps, resulting in a significantly lower sampling cost. Specifically, at inference time, \our{} is \textbf{32x} and \textbf{11x} more faster than CDVAE and DiffCSP, respectively, on the Perov-5 dataset, and \textbf{45x} and \textbf{6x} more faster on the MP-20 dataset. Overall, these results highlight the effectiveness of \our{} in time- and resource-constrained scenarios. Like other baseline diffusion models, \our{} is capable of generating stable and valid materials. However, by significantly reducing the sampling time per stable material, it enables users to generate more stable materials within a limited time budget compared to other models.\\
To further investigate how well \our{} generates low-energy structures compared to baseline models, we conducted an experiment using the MP-20 dataset. We plotted the histogram of the computed \(E^{hull}\) distribution for relaxed structures across different methods in Figure \ref{fig:hist}. Our observations indicate that, on average, \our{} produces a higher proportion of low-energy structures than CDVAE while being competitive with DiffCSP. Thus, the structures generated by \our{} are, on average, more stable than those generated by CDVAE and comparable in stability to those generated by DiffCSP.

\section{Conclusion}
\label{conclusion}
In this work, we focus on generating stable crystal materials within a time- and resource-constrained setup by exploring latent diffusion models. We introduce \our{}, a novel latent diffusion model that operates in a lower-dimensional, smooth latent space, making it more efficient in terms of time and resource consumption. Extensive experiments on benchmark generative tasks using two popular datasets demonstrate that \our{} produces novel materials with comparable validity and stability to existing methods while being significantly faster during both training and inference. Additionally, \our{} generates a higher proportion of lower-energy structures compared to baseline models.

{\small
\bibliographystyle{plainnat}
\bibliography{references}
}

\clearpage
\appendix
\section*{Appendix}

\section{More Related Works}
\label{appendix_related}

\subsection{Crystal Representation Learning}
\label{appendix_related_repr}

\paragraph{Graph neural architectures.}
Graph neural network (GNN) based approaches have emerged as a powerful paradigm for learning robust representations of crystalline materials, enabling fast and accurate property prediction. CGCNN~\cite{xie2018crystal} was the first such model: it represents a 3D crystal structure as an undirected, weighted multi-edge graph and applies graph convolutions directly on it. Following CGCNN, a large body of work~\cite{chen2019graph,choudhary2021atomistic,das2023crysmmnet,louis2020graph,Wolverton2020,schmidt2021crystal} proposed GNN variants that incorporate richer geometric information such as bond angles, multi-modal inputs, and attention mechanisms. More recently, transformer-based architectures have been tailored to periodic structures. Matformer~\cite{yan2022periodic} learns periodic graph representations and matches or marginally exceeds prior state-of-the-art accuracy while being substantially faster. PotNet~\cite{lin2023efficient} approximates complete interatomic potentials over infinite periodic summations, ComFormer~\cite{yan2024complete} builds complete and efficient SE(3)-invariant and equivariant crystal graphs, and Crystalformer~\cite{taniai2024crystalformer} introduces infinitely connected attention to encode the full periodic structure.

\paragraph{Pre-training and data-efficient learning.}
A key bottleneck for these models is the scarcity of labeled data, as DFT-computed or experimentally measured properties are expensive to obtain for many targets. To mitigate this, CrysXPP~\cite{das2022crysxpp} transfers knowledge from an autoencoder trained on large volumes of unlabeled crystal graphs, while CrysGNN~\cite{das2023crysgnn} pre-trains a GNN with node- and graph-level self-supervised objectives and distills the learned knowledge into arbitrary property predictors. Crystal Twins~\cite{magar2022crystal} adopts contrastive and redundancy-reduction objectives for self-supervised pre-training. Inspired by the success of generative pre-training, diffusion-based frameworks such as CrysDiff~\cite{song2024diffusion} and DPF~\cite{shen2025denoising} pre-train encoders by denoising atom types, fractional coordinates, and lattice parameters directly in the structure space, which requires separate noise processes and complex denoising architectures for each modality. CrysLDNet~\cite{mukherjee2026latent} addresses this by moving the diffusion process to a learned latent space: a VAE encoder maps crystal structures into a smooth, compact latent space in which a latent diffusion model is trained, encouraging the graph encoder to capture structural and chemical semantics from large-scale unlabeled data before fine-tuning on downstream properties. Orthogonal to full fine-tuning, CrysPrompt~\cite{mukherjee2026crysprompt} proposes a lightweight, model-agnostic soft prompt learning framework that injects node-level and crystal-system-aware graph-level prompts into a frozen backbone, improving over fine-tuned pre-trained models and enabling cross-property knowledge transfer for properties with limited labels.

\subsection{Diffusion and Flow-based Generative Models}
\label{appendix_related_diffusion}

The fundamental idea of diffusion models, initially proposed by~\citet{sohl2015deep}, is to gradually corrupt data with noise and train a neural network to recover the data from noise. This idea was subsequently developed along two broad lines: 1) \textit{score matching networks}~\citep{song2019generative,song2020improved} and 2) \textit{denoising diffusion probabilistic models (DDPMs)}~\citep{ho2020denoising}. Diffusion models have since become a powerful family of deep generative models, achieving remarkable performance across applications such as image synthesis~\citep{dhariwal2021diffusion}, molecular conformer generation~\citep{shi2021learning,xu2022geodiff}, molecular graph generation~\citep{liu2021graphebm}, and protein modeling~\citep{wu2021se,luo2022antigen}.

\paragraph{Latent diffusion.}
Several studies have developed latent diffusion models (LDMs), which run the diffusion process in the compressed latent space of a pre-trained autoencoder, with promising results in image generation~\citep{vahdat2021score}, point clouds~\citep{vahdat2022lion}, and text generation~\citep{li2022diffusion}. Among them, Stable Diffusion~\citep{rombach2022high} demonstrates remarkably realistic text-guided image generation, and its success motivates latent-space generative modeling for scientific data such as crystals, where heterogeneous discrete and continuous modalities can be unified in a single latent representation.

\paragraph{Flow matching.}
Flow matching~\citep{lipman2023flow} offers a simulation-free alternative that regresses a time-dependent vector field transporting a simple prior to the data distribution, and has been extended to Riemannian manifolds~\citep{chen2024flow}. Since fractional coordinates lie on a torus and lattices obey geometric constraints, Riemannian flow matching is a natural fit for crystalline materials and typically requires far fewer integration steps than diffusion sampling.

\subsection{Crystal Material Generation}
\label{appendix_related_generation}

\paragraph{Early approaches.}
Early efforts to create novel periodic materials concentrated on generating atomic compositions while largely neglecting 3D structure. With the advancement of deep generative models, subsequent work used VAEs and GANs to generate 3D periodic structures, either representing materials as three-dimensional voxel images~\citep{court20203,hoffmann2019data,long2021constrained,noh2019inverse} from which atom types, coordinates, and lattices are decoded, or directly encoding structures as embedding vectors~\citep{kim2020generative,ren2020inverse,zhao2021high}. However, these models neither explicitly promote stability of the generated structures nor are invariant to Euclidean and periodic transformations.

\paragraph{Equivariant diffusion and flow models.}
Equivariant diffusion models opened a promising trajectory for generating novel 3D periodic structures. CDVAE~\citep{xie2021crystal} was the first work to integrate a variational autoencoder with a score-based decoder that operates directly on atomic coordinates, using an equivariant GNN to ensure Euclidean and periodic invariance. Subsequently, numerous studies~\citep{luo2023towards,jiao2023crystal,zeni2023mattergen,jiao2024space,yang2023scalable,lin2024equivariant} used diffusion models to learn the joint distribution of atom types, fractional coordinates, and lattices, enabling generation of stable periodic structures; among them, DiffCSP++~\citep{jiao2024space} additionally enforces space-group constraints, and MatterGen~\citep{zeni2023mattergen} supports fine-tuning towards target chemistry, symmetry, and properties. FlowMM~\citep{miller2024flowmm} adapts Riemannian flow matching to crystals, matching diffusion-based performance with substantially fewer integration steps.

\paragraph{Latent and unified generative models.}
Following the success of LDMs, recent work performs crystal generation in a learned latent space. ADiT~\citep{joshi2025allatom} learns a shared latent space for molecules and materials and trains a single diffusion transformer to generate both.

\paragraph{Language-model-based generation.}
A parallel line of work treats crystals as text. CrystaLLM~\citep{antunes2024crystal} trains an autoregressive language model on CIF files, while \citet{gruver2024finetuned} show that fine-tuning a pre-trained LLaMA-2 model on string-encoded crystals yields a high rate of stable materials and naturally supports infilling and text-conditional generation. CrystalFormer~\citep{cao2024space} autoregressively generates Wyckoff positions with space-group information built in, and CrysText~\citep{mohanty2025crystext} generates CIFs from compositional and space-group prompts using supervised fine-tuning and reinforcement learning. While LLMs capture discrete chemistry and symmetry well, they handle continuous coordinates and lattice parameters as discretized tokens, which limits structural precision.

\paragraph{Hybrid LLM--diffusion models.}
To combine the complementary strengths of the two paradigms, recent hybrid frameworks use LLMs to propose an initial structure that is subsequently refined by a continuous generative model. FlowLLM~\citep{sriram2024flowllm} uses a fine-tuned LLM as the base distribution for Riemannian flow matching. CrysLLMGen~\citep{khastagir2025llm} employs a two-stage sampling process in which a LoRA-fine-tuned LLM first generates an intermediate representation of atom types, coordinates, and lattice, after which a pre-trained equivariant diffusion model refines the continuous coordinates and lattice parameters while preserving the LLM-predicted composition. This yields more stable and novel materials than either LLM-only or denoising-only baselines and retains the LLM's ability to follow user-defined conditional prompts.

\paragraph{Text- and property-conditioned generation.}
Beyond unconditional generation, inverse design requires steering generation towards desired properties or descriptions. TGDMat~\citep{das2025periodic} conditions a joint equivariant diffusion model on textual descriptions of the target material at every denoising step, and Chemeleon~\citep{park2025exploration} aligns text and structure embeddings via contrastive learning to guide crystal diffusion. Text-conditional LLM approaches~\citep{gruver2024finetuned,mohanty2025crystext} and hybrid models such as CrysLLMGen~\citep{khastagir2025llm} further allow flexible natural-language constraints, such as composition, space group, or crystal system, to be specified directly in the prompt.

\section{Invariances in Crystal Structure}
\label{appendix_prelim_inv}
The basic idea of using generative models for crystal generation is to learn the underlying data distribution of material structure $p(\textbf{\textit{M}})$. Since crystal materials satisfy physical symmetry properties ~\cite{dresselhaus2007group,zee2016group}, one of the major challenges here is the learned distribution must satisfy periodic E(3) invariance i.e. invariance to permutation, translation, rotation, and periodic transformations.
\begin{itemize}
    \item \textbf{\textit{Permutation Invariance :}} If we permute the indices of constituent atoms it will not change the material. Formally, given any material $ \textbf{\textit{M}}=(\textbf{\textit{A}},\textbf{\textit{X}},\textbf{\textit{L}})$, using any permutation matrix $\mathbf{P}$ if we permute $\textbf{\textit{A}}$ and $\textbf{\textit{X}}$ as $\mathbf{P}(\textbf{\textit{A}})$ and $\mathbf{P}(\textbf{\textit{X}})$, then new material $\textbf{\textit{M}}_\textbf{\textit{P}}=(\mathbf{P}(\textbf{\textit{A}}),\mathbf{P}(\textbf{\textit{X}}),\textbf{\textit{L}})$ will remains unchanged. Hence the underlying distribution is also the same i.e $p( \textbf{\textit{M}}) = p(\textbf{\textit{M}}_\textbf{\textit{P}})$.
    
    \item \textbf{\textit{Translation Invariance :}} If we translate the atom coordinates by a random vector it will not change the structure of the material. Formally, given any material $ \textbf{\textit{M}}=(\textbf{\textit{A}},\textbf{\textit{X}},\textbf{\textit{L}})$, if we translate $\textbf{\textit{X}}$ by an arbitrary translation vector $\mathbf{u} \in \mathbb{R}^{3}$, new generated material $\textbf{\textit{M}}_\textbf{\textit{P}} =(\textbf{\textit{A}},\textbf{\textit{X}}+\mathbf{u}\mathbf{1}^T,\textbf{\textit{L}})$ will be the same as $ \mathbf{\textit{M}}$. Hence $p( \textbf{\textit{M}}) = p(\textbf{\textit{M}}_\textbf{\textit{T}})$ must satisfy.
    
    \item \textbf{\textit{Rotational Invariance :}} If we rotate the atom coordinates and lattice matrix, the material remains unchanged. Formally, using any orthogonal rotational matrix $\mathbf{Q} \in R^{3 \times 3}$ (satisfying $\mathbf{Q}^T\mathbf{Q}= \mathbf{I}$), if we rotate $\textbf{\textit{X}}$ and $\textbf{\textit{L}}$ of any material $ \textbf{\textit{M}}$ and generate new $\textbf{\textit{M}}_\textbf{\textit{R}}=(\textbf{\textit{A}},\textbf{\textit{QX}},\textbf{\textit{QL}})$, then actually different representations of the same material. Hence $p( \textbf{\textit{M}}) = p(\textbf{\textit{M}}_\textbf{\textit{R}})$ must satisfy.
    
    \item \textbf{\textit{Periodic Invariance :}} Finally, since the atoms in the unit cell can periodically repeat itself infinite times along the lattice vector, there can be many choices of unit cells and coordinate matrices representing the same material. Formally, given coordinates $\textbf{\textit{X}}$, after applying periodic transformation using random matrix $\textbf{\textit{K}} \in R^{n \times 3}$, new coordinates $\mathbf{X'}=\textbf{\textit{X}}+\textbf{\textit{K}}\textbf{\textit{L}}$ are periodically equivalent. Hence $\mathbf{\textit{M}}=(\textbf{\textit{A}},\textbf{\textit{X}},\textbf{\textit{L}})$ and $\textbf{M'}=(\textbf{\textit{A}},\mathbf{X'},\textbf{\textit{L}})$ are same material and $p( \textbf{\textit{M}}) = p(\textbf{M'})$ must hold.
\end{itemize}

\section{Denoising Network Architecture: CSPNet}
\label{app_cspnet}
For the backbone network in the backward diffusion process of \our{}, we extend the CSPNet architecture~\cite{jiao2023crystal}. CSPNet is based on EGNN~\cite{satorras2021n}, ensuring periodic E(3) invariance for periodic crystal structures. At the $k^{th}$ layer of message passing, the Equivariant Graph Convolutional Layer (EGCL) takes as input the atom embeddings $\textbf{\textit{h}}^{k}=[\textbf{\textit{h}}^{k}_1,\textbf{\textit{h}}^{k}_2,...,\textbf{\textit{h}}^{k}_N]$, atom coordinates $\textbf{\textit{x}}^{k}=[\textbf{\textit{x}}^{k}_1,\textbf{\textit{x}}^{k}_2,...,\textbf{\textit{x}}^{k}_N]$, and the lattice matrix \textbf{\textit{L}}, and outputs a transformed set of atom embeddings $\textbf{\textit{h}}^{k+1}$. Formally, the message passing operation at the $k^{th}$ layer is defined as follows:
\begin{equation}
\label{eq:appendix_msg_pass_1}
    \textbf{\textit{m}}_{i,j} = \rho_m \{\textbf{\textit{h}}^{k}_i, \ \textbf{\textit{h}}^{k}_j, \ \textbf{\textit{L}}^T\textbf{\textit{L}}, \ 
    \psi_{FT}(\textbf{\textit{x}}^{k}_i - \textbf{\textit{x}}^{k}_j ) \} ; \;
\end{equation}
\begin{equation}
\label{eq:appendix_msg_pass_2}
    \textbf{\textit{h}}^{k+1}_{i} = \textbf{\textit{h}}^{k}_{i} + \rho_h \{ \textbf{\textit{h}}^{k}_{i}, \textbf{\textit{m}}_{i} \}
\end{equation}
Where $\textbf{\textit{m}}_{i} = \sum_{j=1}^{N} \textbf{\textit{m}}_{i,j}$, $\rho_m$ and $\rho_h$ are multi-layer perceptrons, and $\psi_{FT}$ is a Fourier Transformation function applied to the relative difference between fractional coordinates $\textbf{\textit{x}}^{k}_i$ and $\textbf{\textit{x}}^{k}_j$. The Fourier Transformation is utilized as it remains invariant to periodic translation and captures various frequency components of relative fractional distances, which are essential for accurate crystal structure modeling. Input atom features $\textbf{\textit{h}}^{0}$ and coordinates $\textbf{\textit{x}}^{0}$ are fed through $\mathcal{K}$ layers of EGCL to produce $\hat{\bm{\epsilon}}_{\textbf{\textit{z}}_{L}}$, $\hat{\bm{\epsilon}}_{\textbf{\textit{z}}_{h}}$ and $\hat{\bm{\epsilon}}_{\textbf{\textit{X}}}$ as follows :
\begin{equation}
    \label{eq:appendix_msg_pass_3}
    \begin{split}
        \hat{\bm{\epsilon}}_{\textbf{\textit{z}}_{L}} = \textbf{\textit{L}} \rho_L (\frac{1}{N} \sum^{i=1}_{N} \textbf{\textit{h}}^{\mathcal{K}}) ; \; \\
        \hat{\bm{\epsilon}}_{\textbf{\textit{z}}_{h}} = \rho_A (\textbf{\textit{h}}^{\mathcal{K}}) ; \; \\
        \hat{\bm{\epsilon}}_{\textbf{\textit{X}}} = \rho_X (\textbf{\textit{h}}^{\mathcal{K}})
    \end{split}
\end{equation}
where $\rho_L, \rho_A, \rho_X$ are multi-layer perceptrons on the final layer embeddings.

\section{Training and Sampling Algorithm for \our{}}
\label{app_algo}
\begin{algorithm}
   \caption{Training Algorithm of \our{}}
   \label{alg:training}
\begin{algorithmic}[1]
   \STATE {\bfseries Input:} Crystal Material $\textbf{\textit{M}} = (\textbf{\textit{A}}, \textbf{\textit{X}}, \textbf{\textit{L}})$, Encoder $\mathcal{E}_\psi$, Decoder $\mathcal{D}_\phi$, and Denosing Network $f_\theta$.
   \STATE {\bfseries \underline{Stage-1: Training VAE}}
   \REPEAT
   \STATE $\bm{\mu}_\textbf{h},\bm{\mu}_\textbf{L} \leftarrow \mathcal{E}_\phi(\textbf{\textit{A}}, \textbf{\textit{X}}, \textbf{\textit{L}})$
   \STATE Sample $\bm{\epsilon}^\textbf{h}, \bm{\epsilon}^\textbf{L} \sim N(\textbf{\textit{0}},\textbf{\textit{I}})$
   \STATE $\textbf{\textit{z}}_h \leftarrow \bm{\mu}_\textbf{h} + \bm{\epsilon}^\textbf{h} \odot \sigma_\textbf{h}$
   \STATE $\textbf{\textit{z}}_L \leftarrow \bm{\mu}_\textbf{L} + \bm{\epsilon}^\textbf{L} \odot \sigma_\textbf{L}$
   \STATE $\tilde{\textbf{\textit{A}}}, \tilde{\textbf{\textit{L}}}  \leftarrow
       \mathcal{D}_\psi(\textbf{\textit{z}}_h, \textbf{\textit{X}}, \textbf{\textit{z}}_L)$

   \STATE $ \mathcal{L}^{\textbf{\textit{A}}}_{recon} = CrossEntropyLoss (\tilde{\textbf{\textit{A}}},\textbf{\textit{A}})$

   \STATE $ \mathcal{L}^{\textbf{\textit{L}}}_{recon} = \lVert  \tilde{\textbf{\textit{L}}} \ - \ \textbf{\textit{L}} {\rVert}^2_2$
   
   \STATE Minimize $\mathcal{L}_{VAE} = \mathcal{L}^{\textbf{\textit{A}}}_{recon} + \mathcal{L}^{\textbf{\textit{L}}}_{recon} + \lambda \mathcal{L}_{reg}$ and update parameters of $\mathcal{E}_\psi$ and $\mathcal{D}_\phi$
   \UNTIL{Converged}\\
   
   \STATE {\bfseries \underline{Stage-2: Training LDM}}
   \REPEAT 
    \STATE Sample $t \sim  \mathcal{U}(\mathbf{0},\mathbf{T})$
    \STATE Sample Noise $\epsilon_\textbf{X}, \bm{\epsilon}_{\textbf{\textit{z}}_{h}}, \bm{\epsilon}_{\textbf{\textit{z}}_{L}} \sim N(0,I)$
    \STATE $\textbf{\textit{z}}_{h,t} = \sqrt{\bar{\alpha}_{t}}\textbf{\textit{z}}_{h,0} +  \sqrt{1-\bar{\alpha}_{t}}\bm{\epsilon}_{\textbf{\textit{z}}_{h}}$
    \STATE $\textbf{\textit{z}}_{L,t} = \sqrt{\bar{\alpha}_{t}}\textbf{\textit{z}}_{L,0} +  \sqrt{1-\bar{\alpha}_{t}}\bm{\epsilon}_{\textbf{\textit{z}}_{L}}$
    \STATE $\textbf{\textit{X}}_{t} = f_w(\textbf{\textit{X}}_{0} + \bm{\sigma_t}\bm{\epsilon}^{x})$
    
    \STATE $\hat{\bm{\epsilon}}_{\textbf{\textit{z}}_{h}},\hat{\bm{\epsilon}}_{\textbf{X}},\hat{\bm{\epsilon}}_{\textbf{\textit{z}}_{L}}  \leftarrow f_\theta(\textbf{\textit{z}}_{h,t},\textbf{X}_{t},\textbf{\textit{z}}_{L,t},t)$

    \STATE $\mathcal{L}_{type} = \lVert  \bm{\epsilon}_{\textbf{\textit{z}}_{h}} \ - \ \hat{\bm{\epsilon}}_{\textbf{\textit{z}}_{h}} {\rVert}^2_2 $
    \STATE $\mathcal{L}_{lattice} = \lVert  \bm{\epsilon}_{\textbf{\textit{z}}_{L}} \ - \ \hat{\bm{\epsilon}}_{\textbf{\textit{z}}_{L}} {\rVert}^2_2 $
    \STATE $\mathcal{L}_{coord} = 
            \lVert  \nabla_{\textbf{\textit{X}}_{t}} \text{log} q(\textbf{\textit{X}}_{t} | \textbf{\textit{X}}_{0}) - \hat{\bm{\epsilon}}_{\textbf{\textit{X}}} {\rVert}^2_2$
    
    \STATE Minimize $\mathcal{L}_{LDM} = \lambda_{L} \mathcal{L}_{lattice} + \lambda_{A} \mathcal{L}_{type} + \lambda_{X} \mathcal{L}_{coord}$ and update parameters of $f_\theta$
    \UNTIL{Converged}
\end{algorithmic}
\end{algorithm}

\begin{algorithm}
\caption{Sampling Algorithm of \our{}}\label{alg:sampling}
\begin{algorithmic}[1]
\STATE $N \sim p(N)$
\STATE Sample $\textbf{\textit{z}}_{h,T},\textbf{\textit{z}}_{L,T} \sim  \mathcal{N}(\mathbf{0},\mathbf{I}),\textbf{\textit{X}}_T \sim  \mathcal{U}(0,1)$

\FOR{$t \gets T$ to $1$}
    \STATE $\bm{\epsilon}_{\textbf{\textit{z}}_h},\epsilon_\textbf{X},\bm{\epsilon}_{\textbf{\textit{z}}_L} \sim N(0,I) /*Sample*/$
    \STATE $\hat{\bm{\epsilon}}_{\textbf{\textit{z}}_{h}},\hat{\bm{\epsilon}}_{\textbf{X}},\hat{\bm{\epsilon}}_{\textbf{\textit{z}}_{L}}  \leftarrow f_\theta(\textbf{\textit{z}}_{h,t},\textbf{X}_{t},\textbf{\textit{z}}_{L,t},t)$
    
    \STATE $\textbf{\textit{z}}_{h,t-1} \leftarrow \frac{1}{\sqrt{\alpha_{t}}} (\textbf{\textit{z}}_{h,t} - \frac{\beta_t}{\sqrt{1-\bar{\alpha}_{t}}}\hat{\bm{\epsilon}}_{\textbf{\textit{z}}_{h}})+\sqrt{\beta_t\frac{1-\bar{\alpha}_{t-1}}{1-\bar{\alpha}_{t}}}\bm{\epsilon}_{\textbf{\textit{z}}_h}$

    \STATE $\textbf{\textit{z}}_{L,t-1} \leftarrow \frac{1}{\sqrt{\alpha_{t}}} (\textbf{\textit{z}}_{L,t} - \frac{\beta_t}{\sqrt{1-\bar{\alpha}_{t}}}\hat{\bm{\epsilon}}_{\textbf{\textit{z}}_{L}})+\sqrt{\beta_t\frac{1-\bar{\alpha}_{t-1}}{1-\bar{\alpha}_{t}}}\bm{\epsilon}_{\textbf{\textit{z}}_L}$
    
    \STATE $\textbf{X}_{t-\frac{1}{2}} \leftarrow w(\textbf{X}_{t}+(\sigma^2_t - \sigma^2_{t-1})\hat{\epsilon}^{\textbf{X}}+\frac{\sigma_{t-1}\sqrt{\sigma^2_t - \sigma^2_{t-1}}}{\sigma_{t}}\bm{\epsilon}^\textbf{X})$
    
    \STATE $_,\hat{\bm{\epsilon}}_{\textbf{X}} \leftarrow f_\theta(\textbf{\textit{z}}_{h,t},\textbf{X}_{t-\frac{1}{2}},\textbf{\textit{z}}_{L,t},t)$
    
    \STATE $\eta_t \leftarrow step\_size * \frac{\sigma_{t-1}}{\sigma_t}$
    \STATE $\textbf{X}_{t-1} \leftarrow w(\textbf{X}_{t-\frac{1}{2}}+\eta_t\hat{\bm{\epsilon}}^{\textbf{X}}+\sqrt{2\eta_t}\bm{\epsilon}^\textbf{X})$   
\ENDFOR
\STATE $\textbf{\textit{A}}, \textbf{\textit{X}}, \textbf{\textit{L}}  = 
       \mathcal{D}_\psi(\textbf{\textit{z}}_{h,0}, \textbf{\textit{X}}_0, \textbf{\textit{z}}_{L,0})$
       \STATE \textbf{return} $\textbf{\textit{A}}, \textbf{\textit{X}}, \textbf{\textit{L}}$
\end{algorithmic}
\end{algorithm}

\end{document}